\documentclass[runningheads]{llncs}
\usepackage[T1]{fontenc}
\usepackage{graphicx,verbatim}
\usepackage{booktabs}   
\usepackage{makecell}   
\usepackage{hyperref}
\usepackage{multirow}
\usepackage{amsmath}
\hypersetup{
    colorlinks=true,
    linkcolor=blue,      
    citecolor=blue,      
    urlcolor=blue,       
    linktoc=all
}

\begin{document}
\title{Do Medical Foundation Models Generalize on\\the African Brain?}

\author{Kaouther Mouheb \and Gonzalo Esteban Mosquera Rojas \and Juancito van Leeuwen \and Stefan Klein \and Esther E. Bron}  
\authorrunning{K. Mouheb et al.}
\institute{Dept. of Radiology \& Nuclear Medicine, Erasmus MC, Rotterdam, The Netherlands \\
    \email{k.mouheb@erasmusmc.nl}}

\maketitle              

\begin{abstract}
Medical foundation models (FMs) are increasingly used for brain MRI analysis. However, their evaluation remains dominated by high-resource datasets, leaving generalization to African cohorts underexplored. We assess whether FMs generalize equally to African and non-African brain MRI data across two tasks: dementia classification using a Nigerian dataset and brain tumor segmentation using BraTS-Africa. We evaluate two generalist FMs (BrainIAC, 3DINO) and two segmentation-specific FMs (MedSAM2, Medical-SAM2) against a from-scratch baseline. For classification, FMs provide limited gains (highest ROC-AUC of 0.86 with BrainIAC), whereas for segmentation they consistently improve performance, reaching up to 0.86 Dice with MedSAM2. Performance differences between African and non-African cohorts are inconsistent and appear more related to dataset size than data origin. These results suggest that FMs do not exhibit an inherent bias against African cohorts, and highlight the limited availability and diversity of African neuroimaging datasets as the main barrier to robust evaluation and deployment.
\end{abstract}

\section{Introduction}

Medical image analysis is shifting from specialized fully-supervised models toward foundation models (FMs), which are trained on large-scale heterogeneous data to learn general-purpose representations. This allows FMs to have strong generalization across anatomies, modalities, and downstream tasks \cite{paschali2025foundation}. Although few FMs are specifically pretrained on brain scans, both brain-specific and general-purpose FMs are increasingly applied to neuroimaging. Currently, available FMs can be broadly divided into two categories: those trained specifically for segmentation, e.g. Med-SAM2 \cite{ma2025medsam2}, and those trained as task-independent feature extractors with self-supervision, such as 3DINO \cite{xu2025generalizable}. On well-curated benchmarks such as the brain tumor segmentation challenge (BraTS) \cite{menze2014multimodal}, FMs like BrainSegFounder \cite{wang2025sam} achieve Dice scores of up to $0.90$, which is on par with specialized models, showing the potential of FMs for neuroimage analysis.

However, the training data of medical FMs is predominantly derived from high-resource regions, particularly North America and Europe \cite{danaei2025state,musa2026systematic}. 
This imbalance raises concerns about the external validity of the reported performance and the ability of medical FMs to generalize when applied to underrepresented groups. African populations in particular remain largely absent from both the pretraining datasets and the evaluation benchmarks of current FMs, introducing a critical gap in current evaluation practices. In the context of brain MRI analysis, recent studies on Sub-Saharan African data demonstrate that models trained on standard datasets often exhibit degraded performance when applied to African cohorts \cite{adewole2025brats}. These findings suggest that strong performance on established benchmarks does not necessarily translate to robust generalization on clinical African brain MRI data. To address this gap, recent efforts have introduced brain MRI datasets that reflect African clinical settings. The BraTS-Africa 2023 challenge released a multi-institutional Sub-Saharan MRI dataset for brain tumor segmentation \cite{adewole2025brats} with heterogeneous acquisition protocols and image quality, and Wogu et al. \cite{wogu2025labeled} released a Nigerian brain MRI dataset for the study of neurodegenerative diseases. Despite these advances, a systematic evaluation of medical FMs on African brain MRI data is still missing, leaving an open question: \textit{Is there a generalization gap between African and non-African populations when applying medical FMs to neuroimaging tasks?}

In this work, we present a comprehensive evaluation of four medical foundation models, spanning segmentation-specific and generalist architectures, on one classification and one segmentation task, using two Sub-Saharan African brain MRI datasets. We benchmark the performance against two high-resource datasets from North America and Europe. Through this study, we (i) quantify how well medical FMs generalize to African brain MRI data relative to high-resource cohorts, and (ii) underscore the importance of inclusive benchmarking for developing equitable medical FMs.

\section{Materials and Methods}
\subsection{Datasets}
\label{sec:datasets}
To assess the generalization of FMs on African neuroimaging data, we used two publicly available datasets originating from Sub-Saharan African clinical centers. We compare the results to two high-resource datasets with the same task.\\
\textbf{Dementia classification}: For the classification task, we use a Nigerian brain MRI dataset \cite{wogu2025labeled} comprising T1-weighted scans from 50 subjects (22 female; age $48.8 \pm 19.9$), including dementia (DE, $n=23$) and healthy control (HC, $n=27$) cases. For comparison, we use OASIS-4, a U.S. clinical cohort. We include all 47 HC subjects and 50 randomly sampled DE cases to obtain a balanced dataset of 97 subjects (50 female; age $72.14 \pm 9.4$). All scans are preprocessed using HD-BET for brain extraction \cite{isensee2019automated}, N4 bias field correction \cite{tustison2010n4itk}, and FSL FLIRT for registration to the MNI152 space \cite{fischer2003flirt}.\\
\textbf{Brain tumor segmentation}: For the segmentation task, we use the BraTS-Africa dataset \cite{adewole2025brats}, comprising 146 multi-parametric MRI studies (T1, T1 with Gadolinium, T2, and FLAIR) from Sub-Saharan African medical centers, including 95 glioma and 51 non-glioma cases. Demographic metadata (age, sex) is not reported. We compare against the Erasmus Glioma Dataset (EGD), a high-resource counterpart from the Netherlands, using a size-matched random subset of 150 cases to control for sample size as a confounder, we refer to this subset as EGD-150. The remaining EGD subjects (n=625) are used to assess how fine-tuning sample size affects the performance of the generalistic FMs. Since BraTS-Africa provides tumor-subtype labels while EGD has only whole-tumor masks, we derive whole-tumor masks for BraTS-Africa by taking the union of its sub-region masks to keep the task consistent. Scans are released already preprocessed (co-registered, skull-stripped, resampled); we apply no further preprocessing.\\
All experiments use Monte-Carlo cross-validation (MCCV) with $5$ random train-test splits stratified by class label, with $50\%$ of the data in each split.

\subsection{Models}
We benchmark four FMs satisfying two selection criteria: (i) the model operates on 3D inputs, either via a volumetric encoder or by propagating predictions across a stack of 2D slices, and (ii) the model is pre-trained on medical data, including MRI. These comprise two generalist FMs, BrainIAC \cite{tak2026generalizable} and 3DINO \cite{xu2025generalizable}, and two segmentation-specific FMs, MedSAM2 \cite{ma2025medsam2} and Medical-SAM2 \cite{zhu2024medical}. Importantly, none of these models is reported to have used the datasets used in this study during pretraining.

\subsection{Methods}
This section details the pipelines applied in this study. For both tasks, we train conventional CNNs from scratch on the data as baselines using DenseNet121 for classification and a 3D U-Net for segmentation. The code and further implementation details are publicly available at \href{https://anonymous.4open.science/r/MFM-African-Brain-78C8}{anonymous.4open.science/r/MFM-African-Brain-78C8}.\\
\textbf{Classification:}
We evaluate the models on the task of classifying DE vs. HC subjects. For segmentation-specific FMs we use the image encoder as the feature extractor. We explore two approaches of adapting FMs for a classification task:
\begin{enumerate}
    \item \textbf{Linear Probing:} The image encoder is frozen and used to extract features from each scan. For models producing volumetric or slice-wise features, we apply global average pooling over the spatial dimensions to obtain a 1D feature vector. A linear classifier is then trained to perform the classification based on these 1D vectors. 
    \item \textbf{Parameter-efficient fine-tuning:} To assess whether moderate adaptation of the encoder can improve performance compared to linear probing, we apply parameter-efficient fine-tuning (PEFT), which updates only a small subset of parameters instead of the full model. Specifically, we use low-rank adaptation (LoRA) \cite{hu2022lora}, which injects trainable low-rank matrices into each transformer block while keeping the pre-trained weights frozen, and jointly train the linear classifier. We apply LoRA to BrainIAC and 3DINO. MedSAM2 and Medical-SAM2 are excluded, as their large input size exceeded the available GPU memory for end-to-end fine-tuning. 
\end{enumerate}
\textbf{Segmentation:} For segmentation, segmentation-specific and generalist models are used in different ways:
\begin{enumerate}
    \item \textbf{Segmentation-specific FMs:} We evaluate the zero-shot performance of SAM-based models in their native promptable setting, without task-specific fine-tuning. Following the prompting scheme described by the model developers, we use the bounding box of the tumor's middle slice as a 2D prompt, which the model's video-style decoder then propagates across slices. We explore two settings: (i) generating a probability map per sequence and averaging them to obtain the final mask, and (ii) using FLAIR only, since it is the most informative sequence for whole-tumor segmentation \cite{van2023combined}.
    \item \textbf{Generalist FMs:} for BrainIAC and 3DINO, which are not promptable segmentation models, we attach a UNETR-style decoder \cite{hatamizadeh2022unetr} on top of the frozen pre-trained encoder and train the decoder end-to-end. Since the models take only 1 channel as input we use the FLAIR sequence in this experiment. For EGD, we first perform the experiments using the size-matched subset EGD-150 (training size = 75 in each fold). In a second experiment, we add the remaining 625 samples to the training samples of each MCCV split to evaluate whether increasing the training set size could improve performance. \\
\end{enumerate}
\textbf{Performance Evaluation:} We use area under the receiver operating characteristic curve (ROC-AUC) for classification, and the Dice score (DSC) for segmentation. Given the small number of splits (n=5), which limits the statistical power of formal hypothesis testing, we adopt a descriptive approach: results are summarized as mean $\pm$ standard deviation across splits, and differences between models (both relative to the from-scratch baseline and pairwise across FMs) are assessed in terms of mean performance differences.

\section{Results}
\subsection{Classification}

\begin{table}[t]
\centering
\caption{ROC-AUC of the dementia classification task (mean$_{\pm\text{std}}$ across the test sets of $5$ MCCV splits).}
\label{tab:classification_results}
\setlength{\tabcolsep}{5pt}
\footnotesize
\providecommand{\val}[2]{$#1_{\pm #2}$}
\newcolumntype{C}{>{\centering\arraybackslash}p{3cm}}
\begin{tabular}{l C C}
\toprule
\textbf{Model} & \textbf{Nigerian Brain} & \textbf{OASIS-4} \\
\midrule
DenseNet121 baseline & \val{0.85}{0.10} & \val{0.80}{0.03} \\
\midrule
\multicolumn{3}{l}{\textit{Linear probing}} \\
BrainIAC     & \val{0.85}{0.06} & \val{0.80}{0.05} \\
3DINO        & \val{0.80}{0.03} & \val{0.83}{0.06} \\
MedSAM2      & \val{0.66}{0.09} & \val{0.73}{0.06} \\
Medical-SAM2 & \val{0.59}{0.07} & \val{0.68}{0.03} \\
\midrule
\multicolumn{3}{l}{\textit{Fine-tuning}} \\
BrainIAC + LoRA  & \val{0.86}{0.05} & \val{0.81}{0.02} \\
3DINO + LoRA     & \val{0.82}{0.05} & \val{0.84}{0.04} \\
\bottomrule
\end{tabular}
\end{table}
Table~\ref{tab:classification_results} reports the ROC-AUC for all models on the classification task.
On the Nigerian Brain dataset, the DenseNet121 baseline achieves a ROC-AUC of $0.85 \pm 0.10$. Using linear probing, BrainIAC obtains the highest average ROC-AUC ($0.85 \pm 0.06$), matching the baseline, followed by 3DINO with $0.80 \pm 0.03$, MedSAM2 with $0.66 \pm 0.09$, and Medical-SAM2 with $0.59 \pm 0.07$. With LoRA fine-tuning, the performance improves slightly with $0.86 \pm 0.05$ for BrainIAC and $0.82 \pm 0.05$ for 3DINO. On OASIS-4, the DenseNet121 baseline achieves a ROC-AUC of $0.80 \pm 0.03$. With linear probing, 3DINO obtains the highest average ROC-AUC at $0.83 \pm 0.06$, followed by BrainIAC at $0.80 \pm 0.05$, MedSAM2 at $0.73 \pm 0.06$, and Medical-SAM2 at $0.68 \pm 0.03$. With LoRA fine-tuning, both models show a slight performance increase where 3DINO reaches $0.84 \pm 0.04$ and BrainIAC reaches $0.81 \pm 0.02$. Since Nigerian Brain and OASIS-4 are different datasets, the ROC-AUC values cannot be compared one-to-one to claim that a model performs strictly better or worse on one versus the other. However, the gap between the two datasets varies considerably between models. For segmentation-specific FMs, the gap is relatively larger (up to 9\%) compared to generalist models. 

\subsection{Segmentation}
\label{sec:results_segmentation}

\begin{table}[t]
\centering
\caption{Tumor segmentation Dice score for segmentation-specific FMs obtained with zero-shot prompting. "Combined" refers to the results obtained by averaging the output probability maps of the four sequences. Results are reported as mean$_{\pm\text{std}}$ across the test sets of $5$ Monte-Carlo splits.}
\label{tab:segmentation_results_SFM}
\setlength{\tabcolsep}{5pt}
\footnotesize
\providecommand{\val}[2]{$#1_{\pm #2}$}
\newcolumntype{C}{>{\centering\arraybackslash}p{2cm}}
\begin{tabular}{l C C | C C}
\toprule
& \multicolumn{2}{c|}{\textbf{BraTS-Africa}} & \multicolumn{2}{c}{\textbf{EGD-150}} \\
\cmidrule(lr){2-3} \cmidrule(lr){4-5}
\textbf{Model} & \textbf{Combined} & \textbf{FLAIR} & \textbf{Combined} & \textbf{FLAIR} \\
\midrule
Baseline (U-Net) & \val{0.21}{0.31} & \val{0.55}{0.09} & \val{0.34}{0.20} & \val{0.62}{0.08} \\
\midrule
MedSAM2      & \val{0.86}{0.01} & \val{0.86}{0.01} & \val{0.82}{0.02} & \val{0.82}{0.01} \\
Medical-SAM2 & \val{0.62}{0.01} & \val{0.57}{0.02} & \val{0.71}{0.01} & \val{0.70}{0.01} \\
\bottomrule
\end{tabular}
\end{table}

\begin{figure}[t]
\centering
\includegraphics[width=\textwidth]{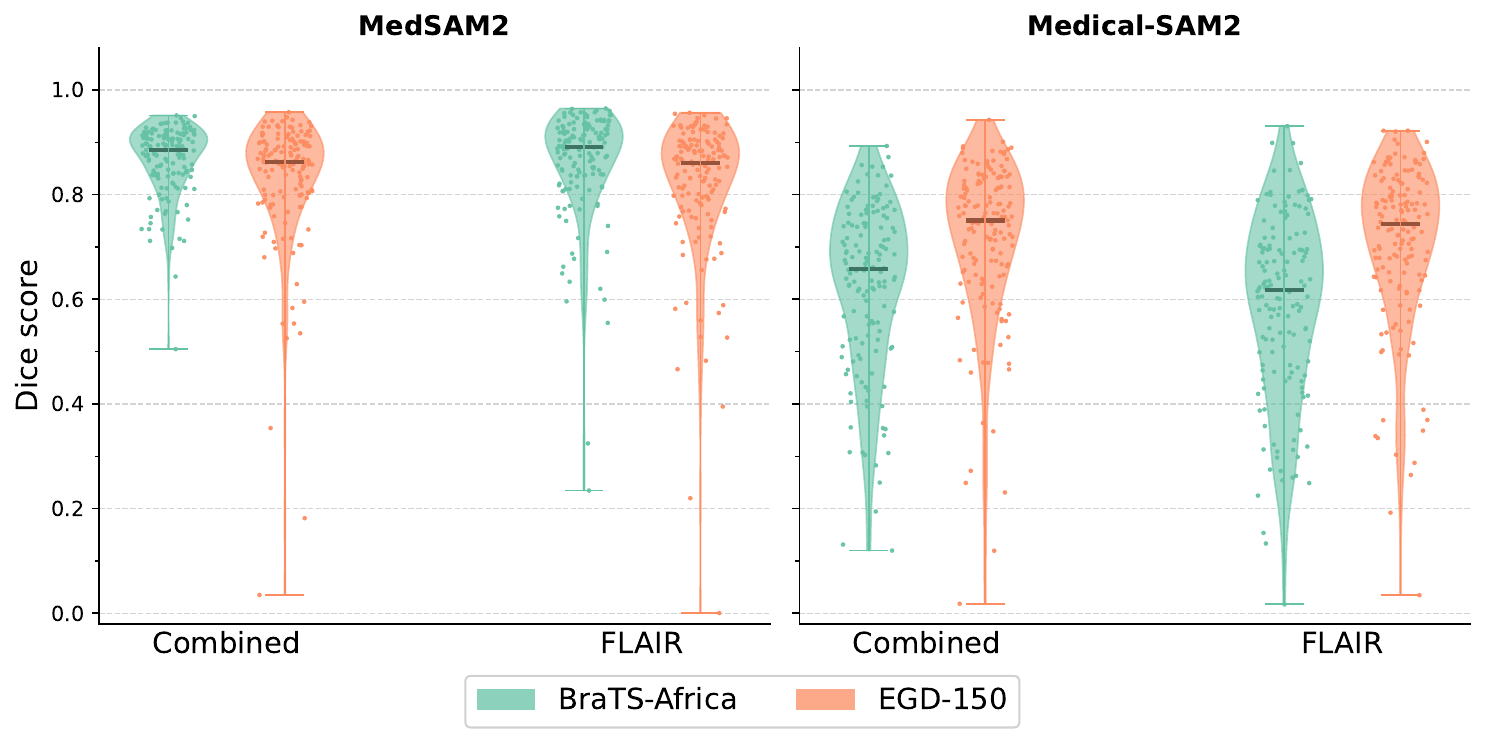}
\caption{Per-case Dice score distributions for zero-shot tumor segmentation on BraTS-Africa ($n=146$) and EGD-150. Combined denotes averaging the output probability maps across all four MRI sequences; FLAIR denotes using the FLAIR sequence only.}
\label{fig:zeroshot_dice}
\end{figure}

Table~\ref{tab:segmentation_results_SFM} summarizes the performance of segmentation-specific FMs in terms of DSC across two settings: Combined (using all modalities) and FLAIR.
In BraTS-Africa, MedSAM2 achieves a DSC of $0.86 \pm 0.01$ in both settings, substantially higher than the U-Net baseline of $0.21 \pm 0.31$ for Combined and $0.55 \pm 0.09$ for FLAIR. Medical-SAM2 achieves $0.62 \pm 0.01$ (Combined) and $0.57 \pm 0.02$ (FLAIR), also higher than the baseline. In EGD-150, MedSAM2 achieves $0.82 \pm 0.02$ for Combined and $0.82 \pm 0.01$ for FLAIR, while Medical-SAM2 achieves $0.71 \pm 0.01$ for Combined and $0.70 \pm 0.01$ for FLAIR, both exceeding the baseline performance of $0.34 \pm 0.20$ (Combined) and $0.62 \pm 0.08$ (FLAIR). 
Both FMs show similar performance between Combined and FLAIR configurations within each dataset, with differences of at most $0.05$ DSC. The U-Net baseline, however, demonstrates pronounced sensitivity to settings, showing substantially lower DSC in Combined versus FLAIR: $0.21 \pm 0.31$ vs.\ $0.55 \pm 0.09$ in BraTS-Africa and $0.34 \pm 0.20$ vs.\ $0.62 \pm 0.08$ in EGD-150.
Figure~\ref{fig:zeroshot_dice} shows zero-shot DSC distributions on the full BraTS-Africa dataset ($n=146$) and EGD-150. For MedSAM2, BraTS-Africa displays tightly concentrated distributions in both settings, with most cases scoring above $0.80$ and few low-scoring outliers, whereas EGD-150 shows wider spread with a longer lower tail toward $0$ despite similar median values. This pattern is reversed for Medical-SAM2: EGD-150 cases are distributed around a higher median ($\sim 0.75$) than BraTS-Africa.

\begin{table}[t]
\centering
\caption{Tumor segmentation Dice score for generalist FMs, using the FLAIR sequence only (mean$_{\pm\text{std}}$ across $5$ Monte-Carlo splits).}
\label{tab:segmentation_results_GFM}
\setlength{\tabcolsep}{5pt}
\footnotesize
\providecommand{\val}[2]{$#1_{\pm #2}$}
\newcolumntype{C}{>{\centering\arraybackslash}p{2.3cm}}
\begin{tabular}{l C C C}
\toprule
\textbf{Model} & \textbf{BraTS-Africa} & \textbf{EGD-150} & \textbf{EGD-large} \\
\midrule
Baseline (U-Net) & \val{0.55}{0.09} & \val{0.62}{0.08} & \val{0.82}{0.02} \\
\midrule
BrainIAC + Decoder & \val{0.63}{0.02} & \val{0.62}{0.03} & \val{0.80}{0.01} \\
3DINO + Decoder  & \val{0.73}{0.02} & \val{0.78}{0.01} & \val{0.82}{0.01} \\
\bottomrule
\end{tabular}
\end{table}

\begin{figure}[t]
\centering
\includegraphics[width=0.85\textwidth]{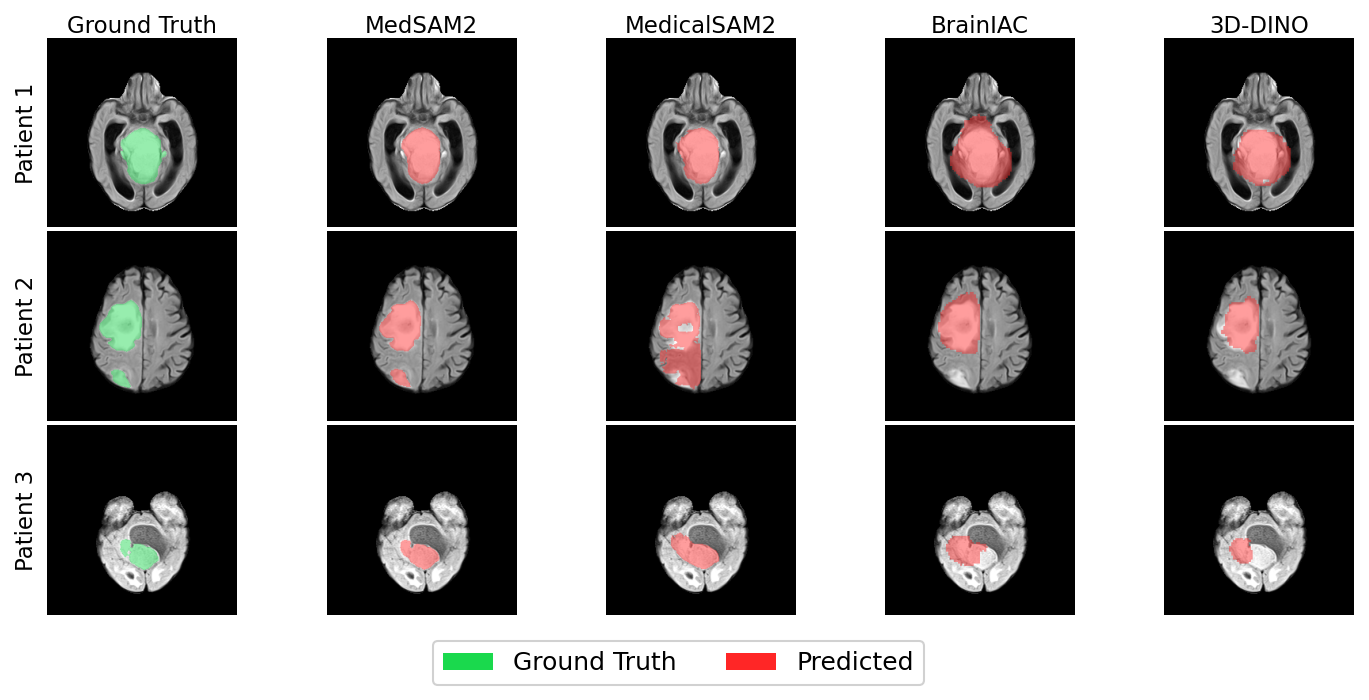}
\caption{Ground truth segmentation and model outputs for three example cases from BraTS-Africa. Results are obtained using the FLAIR sequence as input.}
\label{fig:tumor_slices}
\end{figure}

The segmentation results for generalist FMs are reported in Table~\ref{tab:segmentation_results_GFM}. On BraTS-Africa, both generalist models outperform the U-Net baseline ($0.55 \pm 0.09$) with BrainIAC reaching $0.63 \pm 0.02$ and 3DINO reaching $0.73 \pm 0.02$. On EGD-150, BrainIAC ($0.62 \pm 0.03$) performs comparably to the baseline ($0.62 \pm 0.08$), while 3DINO ($0.78 \pm 0.01$) outperforms it. Comparing the two datasets in the size-matched setting, the performance is similar between BraTS-Africa and EGD-150 for all three models, with differences of at most $0.05$ DSC. With the larger EGD training set, all models improve substantially, reaching $0.80 \pm 0.01$ for BrainIAC and $0.82 \pm 0.01$ for 3DINO. In this setting, both models perform comparably to the from-scratch baseline ($0.82 \pm 0.02$).
Qualitative results on BraTS-Africa (Fig.~\ref{fig:tumor_slices}) show that all four models generally localize the tumor region correctly, but the predictions tend to be less accurate compared to the ground-truth. MedSAM2 produces the closest match to the ground-truth lesion boundary among the four models across the three cases. 

\section{Discussion}
\label{sec:discussion}
In this work, we evaluated whether the generalization of medical foundation models extends equally to African and non-African neuroimaging populations, across one classification and one segmentation task.

When assessing the benefit of FMs over training from scratch, we observed different patterns across tasks. In classification, FMs generally do not offer large improvements compared to training from scratch, whereas in segmentation all FMs outperformed the U-Net baseline. This suggests that pre-trained features transfer well to spatial localization but less to whole-volume classification of subtle pathologies such as dementia. This conclusion holds equally for non-African data, where FMs offer more gains in the segmentation task than in classification. Previous benchmarks have shown similar results; for instance, Li et al. found that FMs do not generalize to tasks with subtle pathologies in X-ray analysis \cite{li2025feature}. LoRA fine-tuning yielded only modest gains over linear probing for both African and non-African data. This could be due to the small size of the fine-tuning data. For both African and non-African data, SAM2-based models showed strong segmentation performance yet performed poorly at classification, possibly because they are optimized for spatial localization rather than for learning globally discriminative semantic representations. Recent work has similarly reported that SAM2 representations may be entangled with localized task-specific cues that limit higher-level semantic understanding \cite{cuttano2026sansa}. 

When comparing African and non-African cohorts under matched sample-size, the performance gap between datasets varied across models, but its direction was not consistent. While in some experiments FMs exhibited higher performance on non-African datasets with up to $0.09$ ROC-AUC and $0.13$ DSC, in most experiments the FMs showed similar patterns between the two populations. This indicates that most of the tested FMs are not inherently biased toward the high-resource population. Notably, increasing the size of the EGD training set substantially improved segmentation performance, suggesting that the advantage of high-resource datasets is driven primarily by sample size rather than by data quality or by their dominant representation in FM pretraining. Further research could explore using explainable AI methods to disentangle performance differences related to data quality from those related to population and demographic differences. Our findings further showed that with a larger training set, FMs no longer outperformed the from-scratch baseline, indicating that their benefit is greatest in limited-data settings. This is a valuable result in this context given the scarcity of African datasets. Promptable segmentation-specific FMs are particularly well suited to this setting, as their strong zero-shot performance allows them to be used without any fine-tuning. Nevertheless, this class of models is highly dependent on prompt quality. In this work, we relied on ground-truth-derived bounding boxes, which may not be available in clinical practice. When some annotated data are available, training a segmentation decoder on top of generalist FMs offers a strong prompt-free alternative.

Some limitations of this study should be considered. First, experiments were conducted on an H100 GPU, which is not representative of typical compute resources in African settings; adapting FMs to lower-resource environments is an important future direction. Second, the African datasets used are small, reflecting the broader scarcity of publicly available African neuroimaging data, which limits the strength of comparative claims and motivated a descriptive analysis. Third, we observed a relatively high classification performance on the Nigerian Brain MRI dataset, although its sample size (n=50) is smaller than that of OASIS-4 (n=97). This performance should be interpreted with caution due to a substantial age gap between the dementia and healthy groups in this dataset, raising the possibility that the models rely in part on age-related features rather than pathology-specific ones. More broadly, African datasets often exhibit design limitations, including demographic confounders and missing metadata, which restrict both subgroup evaluation and the interpretability of model behavior. 

In conclusion, a more rigorous evaluation of the true value of FMs still requires addressing the data limitations we identified in this work, through larger, better-annotated, and more equitably designed African neuroimaging datasets. However, this study highlights the potential of medical FMs to generalize effectively to African populations, with the largest gains observed precisely in the low-data regime that characterizes the African medical imaging landscape.

%
%
\bibliographystyle{splncs04}
\bibliography{mybibliography}
\end{document}